\documentclass{article}
\usepackage{ijcai20}

\usepackage{times}
\usepackage{soul}
\usepackage{url}
\usepackage[hidelinks]{hyperref}
\usepackage[utf8]{inputenc}
\usepackage[small]{caption}
\usepackage{graphicx}
\usepackage{amsmath}
\usepackage{amsthm}
\usepackage{booktabs}
\usepackage{algorithm}
\usepackage{amsmath,amsfonts,bm}

\def\eqref#1{equation~\ref{#1}}

\def\1{\bm{1}}

\def\rd{{\textnormal{d}}}

\def\rs{{\textnormal{s}}}

\def\rvh{{\mathbf{h}}}

\def\rvx{{\mathbf{x}}}

\def\vtheta{{\bm{\theta}}}

\DeclareMathAlphabet{\mathsfit}{\encodingdefault}{\sfdefault}{m}{sl}
\SetMathAlphabet{\mathsfit}{bold}{\encodingdefault}{\sfdefault}{bx}{n}

\usepackage{amsmath}
\usepackage{amssymb}
\usepackage{graphicx}
\usepackage{multirow}
\usepackage{multicol}
\usepackage{subfig}
\usepackage{url}
\usepackage{paralist}
\usepackage{latexsym}
\usepackage{color}
\usepackage{xcolor}
\usepackage{booktabs}
\usepackage{tabularx}
\usepackage{verbatim}
\usepackage{algorithm}
\usepackage{algpseudocode}
\usepackage{CJK}

\newcommand\newcite[1]{{\citeauthor{#1}~[\citeyear{#1}}]}

\title{Towards Making the Most of Context in Neural Machine Translation}

\author{
Zaixiang Zheng$^{1}$\footnote{Equal contribution. This work was done when Zaixiang was visiting at the University of Edinburgh.}
\and
Xiang Yue$^{1*}$\and
Shujian Huang$^{1}$\and
Jiajun Chen$^{1}$\And
Alexandra Birch$^2$
\affiliations
$^1$National Key Laboratory for Novel Software Technology, Nanjing University\\
$^2$ILCC, School of Informatics, University of Edinburgh
\emails
\{zhengzx,xiangyue\}@smail.nju.edu.cn,
\{huangsj,chenjj\}@nju.edu.cn,
a.birch@ed.ac.uk
}

\begin{document}

\maketitle

\begin{abstract}

Document-level machine translation manages to outperform sentence level models by a small margin, but have failed to be widely adopted.
We argue that previous research did not make a clear use of the global context, and propose a new document-level NMT framework that deliberately models the local context of each sentence with the awareness of the global context of the document in both source and target languages. 
We specifically design the model to be able to deal with documents containing any number of sentences, including single sentences. This unified approach allows our model to be trained elegantly on standard datasets without needing to train on sentence and document level data separately.
Experimental results demonstrate that our model outperforms Transformer baselines and previous document-level NMT models with substantial margins of up to 2.1 BLEU on state-of-the-art baselines. We also provide analyses which show the benefit of context far beyond the neighboring two or three sentences, which previous studies have typically incorporated.\footnote{Code was released at \url{https://github.com/Blickwinkel1107/making-the-most-of-context-nmt}}

\end{abstract}

\section{Introduction}
\label{sec:intro}

Recent studies suggest that neural machine translation (NMT)~\cite{sutskever2014sequence,bahdanau2014neural,Vaswani2017Attention} has achieved human parity, especially on resource-rich language pairs~\cite{hassan2018achieving}.
However, standard NMT systems are designed for sentence-level translation, which cannot consider the dependencies among sentences and translate entire documents.
To address the above challenge, various document-level NMT models, viz., context-aware models, are proposed to leverage context beyond a single sentence~\cite{wang2017exploiting,miculicich2018document,zhang2018improving,yang2019enhancing} and have achieved substantial improvements over their context-agnostic counterparts.

Figure \ref{fig:ctx-aware} briefly illustrates typical context-aware models, where the source and/or target document contexts are regarded as an additional input stream parallel to the current sentence, and incorporated into each layer of encoder and/or decoder~\cite{zhang2018improving,tan2019hierarchical}. 
More specifically, the representation of each word in the current sentence is a deep hybrid of both \textit{global} document context and \textit{local} sentence context in every layer.
We notice that these hybrid encoding approaches have two main weaknesses:

\begin{figure}[t]
  \centering
  \includegraphics[width=0.49\textwidth]{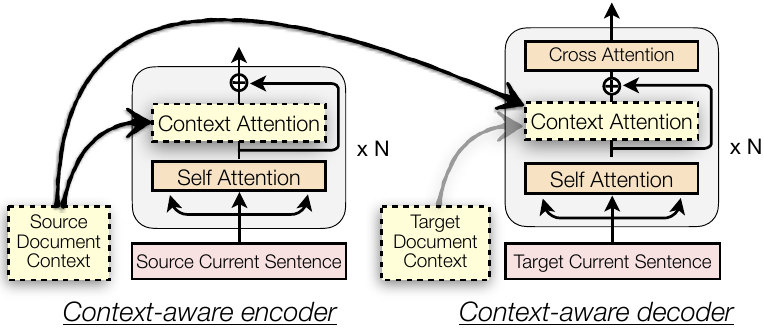}
  \caption{Illustration of typical Transformer-based context-aware approaches (some of them do not consider target context (grey line)).}
  \label{fig:ctx-aware}
\end{figure}

    

\begin{compactitem}
    \item \textit{Models are context-aware, but do not fully exploit the context}.
    The deep hybrid makes the model more sensitive to noise in the context, especially when the context is enlarged.
    This could explain why previous studies show that enlarging context leads to performance degradation.
    Therefore, these approaches have not taken the best advantage of the entire document context.
    
    \item \textit{Models translate documents, but cannot translate single sentences}.  
    Because the deep hybrid requires global document context as additional input, these models are no longer compatible with sentence-level translation based on the solely local sentence context. 
    As a result, these approaches usually translate poorly for single sentence documents without document-level context.
\end{compactitem}

In this paper, we mitigate the aforementioned two weaknesses by designing a general-purpose NMT architecture which can fully exploit the context in documents of arbitrary number of sentences.
To avoid the deep hybrid, our architecture balances \textit{local context} and \textit{global context} in a more deliberate way.
More specifically, our architecture independently encodes local context in the source sentence, instead of mixing it with global context from the beginning so it is robust to when the global context is large and noisy. Furthermore our architecture translates in a sentence-by-sentence manner with access to the partially generated document translation as the target global context which allows the local context to govern the translation process for single-sentence documents.




We highlight our contributions in three aspects:



\begin{compactitem}
    \item We propose a new NMT framework that is able to deal with documents containing any number of sentences, including single-sentence documents, making training and deployment simpler and more flexible. 
    
    \item We conduct experiments on four document-level translation benchmark datasets, which show that the proposed unified approach outperforms Transformer baselines and previous state-of-the-art document-level NMT models both for sentence-level and document-level translation. 
    
    \item Based on thorough analyses, we demonstrate that the document context really matters; and the more context provided, the better our model translates. This finding is in contrast  to the prevailing consensus that a wider context deteriorates translation quality.
\end{compactitem}




\section{Related Work}

Context beyond the current sentence is crucial for machine translation. \newcite{bawden2018evaluating}, \newcite{laubli2018has}, \newcite{mueller2018alarge}, \newcite{voita2018context} and \newcite{voita2019good} show that without access to the document-level context, NMT is likely to fail to maintain lexical, tense, deixis and ellipsis consistencies, resolve anaphoric pronouns and other discourse characteristics, and propose corresponding testsets for evaluating discourse phenomena in NMT. 


Most of the current document-level NMT models can be classified into two main categories, context-aware model, and post-processing model. The post-processing models introduce an additional module that learns to refine the translations produced by context-agnostic NMT systems to be more discourse coherence~\cite{xiong2019modeling,voita2019context}. 
While this kind of approach is easy to deploy, the two-stage generation process may result in error accumulation. 

In this paper, we pay attention mainly to context-aware models, while post-processing approaches can be incorporated with and facilitate any NMT architectures.
\newcite{tiedemann2017neural} and \newcite{junczys2019microsoft} use the concatenation of multiple sentences (usually a small number of preceding sentences) as NMT's input/output.
Going beyond simple concatenation, \newcite{jean2017does} introduce a separate context encoder for a few previous source sentences. \newcite{wang2017exploiting} includes a hierarchical RNN to summarize source context. Other approaches using a dynamic memory to store representations of previously translated contents~\cite{tu2018learning,kuang2018modeling,maruf2018document}. \newcite{miculicich2018document}, \newcite{zhang2018improving}, \newcite{yang2019enhancing}, \newcite{maruf2019selective} and \newcite{tan2019hierarchical} extend context-aware model to Transformer architecture with additional context related modules. 

While claiming that modeling the whole document is not necessary, these models only take into account a few surrounding sentences~\cite{maruf2018document,miculicich2018document,zhang2018improving,yang2019enhancing}, or even only monolingual context~\cite{zhang2018improving,yang2019enhancing,tan2019hierarchical}, which is not necessarily sufficient to translate a document.
On the contrary, our model can consider the entire arbitrary long document and simultaneously exploit contexts in both source and target languages.
Furthermore, most of these document-level models cannot be applied to sentence-level translation, lacking both simplicity and flexibility in practice. They rely on variants of components specifically designed for document context (e.g., encoder/decoder-to-context attention embedded in all layers~\cite{zhang2018improving,miculicich2018document,tan2019hierarchical}), being limited to the scenario where the document context must be the additional input stream. Thanks to our general-purpose modeling, the proposed model manages to do general translation regardless of the number of sentences of the input text.

\begin{figure*}[t]
  \centering
  \includegraphics[width=0.99\textwidth]{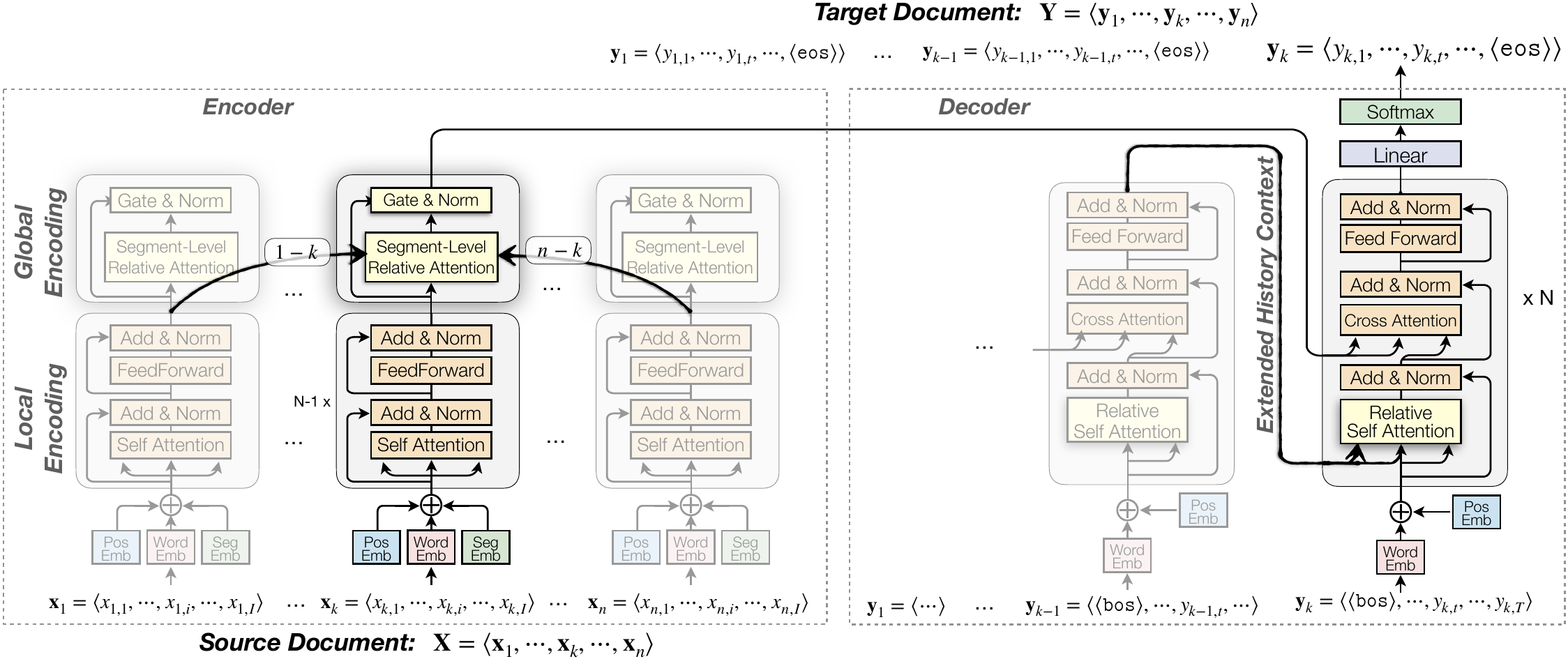}
  \caption{Illustration of the proposed model. The local encoding is complete and independent, which also allows context-agnostic generation.   }
  
  \label{fig:model}
\end{figure*}

\section{Background}
\paragraph{Sentence-level NMT}
Standard NMT models usually model sentence-level translation (\textsc{SentNmt}) within an \textit{encoder-decoder} framework~\cite{bahdanau2014neural}. Here \textsc{SentNmt} models aim to maximize the conditional log-likelihood $\log p(y|x;\theta)$ over a target sentence ${y}=\langle y_1, \dots, y_T\rangle$ given a source sentence ${x}=\langle x_1, \dots, x_I\rangle$ from abundant parallel bilingual data $\mathcal{D}_{\rs} = \{x^{(m)}, y^{(m)}\}_{m=1}^M$ of i.i.d observations: $\mathcal{L}(\mathcal{D}_{\rs};\theta) = \sum_{m=1}^{M} \log p(y^{(m)}|x^{(m)};\theta) $.

\paragraph{Document-level NMT}
Given a document-level parallel dataset $\mathcal{D}_{\rd} = \{X^{(m)}, Y^{(m)}\}_{m=1}^M$, where $X^{(m)} = \langle x^{(m)}_k \rangle_{k=1}^n$ is a source document containing $n$ sentences while $Y^{(m)} = \langle y^{(m)}_k \rangle_{k=1}^n$ is a target document with $n$ sentences, the training criterion for document-level NMT model (\textsc{DocNmt}) is to maximize the conditional log-likelihood over the pairs of document translation sentence by sentence by:
\begin{align}
  \mathcal{L}(\mathcal{D}_{\rd};\theta) &=   \sum_{m=1}^M \log p(Y^{(m)}|X^{(m)};\theta) \nonumber \\
  &=  \sum_{m=1}^M \sum_{k=1}^{n} \log p(y^{(m)}_k|y^{(m)}_{<k},x^{(m)}_{k},x^{(m)}_{-k};\theta) \nonumber
\end{align}
where $y^{(m)}_{<k}$ denotes the history translated sentences prior to $y^{(m)}_{k}$, while $x^{(m)}_{-k}$ means the rest of the source sentences other than the current $k$-th source sentence $x^{(m)}_{k}$.

\section{Approach}



By the definition of local and global contexts, general translation can be seen as a hierarchical natural language understanding and generation problem based on local and global contexts.
Accordingly, we propose a general-purpose architecture to exploit context machine translation to a better extent.
Figure \ref{fig:model} illustrates the idea of our proposed architecture:  
\begin{compactitem}
    \item Given a source document, the encoder builds local context for each individual sentence (local encoding) and then retrieves global context from the entire source document to understand the inter-sentential dependencies (global encoding) and form hybrid contextual representations (context fusion). 
    For single sentence generation, the global encoding will be dynamically disabled and the local context can directly flow through to the decoder to dominate translation. (Section \ref{sec:encoder}) 
    
    \item Once the local and global understanding of the source document is constructed, the decoder generates target document by sentence basis, based on source representations of the current sentence as well as target global context from previous translated history and local context from the partial translation so far. (Section \ref{sec:decoder})
\end{compactitem}

This general-purpose modeling allows the proposed model to fully utilize bilingual and entire document context and go beyond the restricted scenario where models must have document context as additional input streams and fail to translate single sentences. 
These two advantages meet our expectation of a unified and general NMT framework.

\subsection{Encoder}
\label{sec:encoder}
\subsubsection{Lexical and Positional Encoding}
The source input will be transformed to lexical and positional representations.
We use word position embedding in Transformer \cite{Vaswani2017Attention} to represent the order of words. Note that we reset word positions for each sentence, i.e., the $i$-th word in each sentence shares the word position embedding $E^w_{i}$. 
Besides, we introduce segment embedding $E^s_{k}$ to represent the $k$-th sentence. 
Therefore, the representation of $i$-th word in $k$-th sentence is given by $\tilde{x}_{k,i} = E[x_{k,i}] + E^s_{k} + E^w_{i}$, where $E[x_{k,i}]$ means word embedding of $x_{k,i}$.

\subsubsection{Local Context Encoding}
We construct the local context for each sentence with a stack of standard transformer layers~\cite{Vaswani2017Attention}.
Given the $k$-th source sentence $x_k$, the local encoder leverages $N-1$ stacked layers to map it into encoded representations.
\begin{align}
    &\mathbf{\hat{h}}^{l}_k = \mathtt{MultiHead}(\mathtt{SelfAttn}(\mathbf{h}^{l-1}_k,\mathbf{h}^{l-1}_k,\mathbf{h}^{l-1}_k)), \nonumber \\
    &\mathbf{h}^{l}_k = \mathtt{LayerNorm}(\mathtt{FeedForward}(\mathbf{\hat{h}}^{l}_k) + \mathbf{\hat{h}}^{l}_k), \nonumber
\end{align}
where $\mathtt{SelfAttn}(\textbf{Q},\textbf{K},\textbf{V})$ denotes self-attention, while $\textbf{Q},\textbf{K},\textbf{V}$ indicate \textit{queries, keys}, and \textit{values}, respectively. $\mathtt{\mathtt{MultiHead}}(\cdot)$ means the attention is performed in a multi-headed fashion~\cite{Vaswani2017Attention}.
We let the input representations $\tilde{\rvx}_k$ to be the $0$-th layer representations $\mathbf{h}^{0}_k$, while we denote the $(N-1)$-th layer of the local encoder as the local context for each sentence, i.e., $\mathbf{h}^{L}_k = \mathbf{h}^{N-1}_k$.

\subsubsection{Global Context Encoding}
We add an additional layer on the top of the local context encoding layers, which retrieves global context from the entire document by a \textit{segment-level relative attention},
and outputs final representations based on hybrid local and global context by \textit{gated context fusion} mechanism.

\paragraph{Segment-level Relative Attention.}
Given the local representations of each sentences, we propose to extend the relative attention \cite{Shaw2018} from token-level to segment-level to model the inter-sentence global context: 
\begin{align}
    &\mathbf{{h}}^{G} = \mathtt{MultiHead}(\mathtt{Seg\text{-}Attn}(\mathbf{h}^{L},\mathbf{h}^{L},\mathbf{h}^{L})), \nonumber
\end{align}
where $\mathtt{Seg\text{-}Attn}(\textbf{Q},\textbf{K},\textbf{V})$ denotes the proposed segment-level relative attention. Let us take $x_{k,i}$ as query as an example, its the contextual representations $z_{k,i}$ by the proposed attention is computed over all words (e.g., $x_{\kappa,j}$) in the document regarding the sentence (segment) they belong to:
\begin{align}
     &z_{k,i}  = \sum_{\kappa=0}^{n} \sum_{j=1}^{|\mathbf{x}_\kappa|} \alpha^{\kappa,j}_{k,i} ( {W}^V  {x_{\kappa,j}} + \gamma^V_{k-\kappa}), \nonumber \\
     &~~~~~~~~~~~~~~\alpha^{\kappa,j}_{k,i} = \mathtt{softmax}(e^{\kappa,j}_{k,i}), \nonumber
\end{align}
where $\alpha^{\kappa,j}_{k,i}$ is the attention weight of $x_{k,i}$ to $x_{\kappa,j}$. The corresponding attention logit $e^{\kappa,j}_{k,i}$ can be computed with respect to relative sentence distance by:
\begin{align}
     e^{\kappa,j}_{k,i} =  {( {W}^Q x_{k,i})({W}^K x_{\kappa,j}  + \gamma^K_{k-\kappa})^\top} / {\sqrt{d_z}}, \label{eq:seg-attn}
\end{align}
where $\gamma^{*}_{k-\kappa}$ is a parameter vector corresponding to the relative distance between the $k$-th and $\kappa$-th sentences, providing inter-sentential clues. $W^{Q}$, $W^{K}$, and $W^{V}$ are linear projection matrices for the queries, keys and values, respectively.
    
\paragraph{Gated Context Fusion.}
After the global context is retrieved, we adopt a gating mechanism to obtain the final encoder representations $\mathbf{{h}}$ by fusing local and global context:
\begin{align}
    &~~~~~~~~~~~~~~~~~~~~~~\mathbf{g} = \sigma (W_g[\mathbf{{h}}^{L}; \mathbf{h}^{G}]), \nonumber \\
    &\mathbf{{h}} = \mathtt{LayerNorm}\big( (1-\mathbf{g}) \odot \mathbf{{h}}^{L} + \mathbf{g} \odot \mathbf{h}^{G} \big), \nonumber
\end{align}
where $W_g$ is a learnable linear transformation. $[\cdot;\cdot]$ denotes concatenation operation. $\sigma(\cdot)$ is sigmoid activation which leads the value of the fusion gate to be between 0 to 1. $\odot$ indicates element-wise multiplication.

\subsection{Decoder}
\label{sec:decoder}
The goal of the decoder is to generate translations sentence by sentence by considering the generated previous sentences as target global context. A natural idea is to store the hidden states of previous target translations and allow the self attentions of the decoder to access to these hidden states as extended history context.

To that purpose, we leverage and extend Transformer-XL~\cite{dai2019transformer} as the decoder. 
Transformer-XL is a novel Transformer variant, which is designed to cache and reuse the previous computed hidden states in the last segment as an extended context, so that long-term dependency information occurs many words back could propagate through the recurrence connections between segments, which just meets our requirement of generating document long text.
We cast each sentence as a ''segment'' in translation tasks and equip the Transformer-XL based decoder with cross-attention to retrieve time-dependent source context for the current sentence.
Formally, given two consecutive sentences, $y_k$ and $y_{k-1}$, the $l$-th layer of our decoder first employs self-attention over the extended history context:
\begin{align}
    &\mathbf{\tilde{s}}^{l-1}_{k} = [\mathtt{SG}(\mathbf{{s}}^{l-1}_{k-1}); \mathbf{s}^{l-1}_{k}], \nonumber \\
    &\mathbf{\bar{s}}^{l}_{k} = \mathtt{MultiHead}(\mathtt{Rel\text{-}
    SelfAttn}(\mathbf{s}^{l-1}_k,\mathbf{\Tilde{s}}^{l-1}_k,\mathbf{\Tilde{s}}^{l-1}_k)), \nonumber \\
    &\mathbf{\bar{s}}^{l}_k = \mathtt{LayerNorm}(\mathbf{\bar{s}}^{l}_k + \mathbf{s}^{l-1}_k), \nonumber 
\end{align}
where the function $\mathtt{SG}(\cdot)$ stands for stop-gradient. $\mathtt{Rel\text{-}SelfAttn}(\textbf{Q},\textbf{K},\textbf{V})$ is a variant of self-attention with word-level relative position encoding. For more specific details, please refer to \cite{dai2019transformer}. After that, the cross-attention module fetching the source context from encoder representation $\rvh_k$ is computed as:
\begin{align}
    &\mathbf{\hat{s}}_k^{l} = \mathtt{MultiHead}(\mathtt{CrossAttn}(\mathbf{\bar{s}}^{l}_k,\mathbf{{h}}_k,\mathbf{{h}}_k)), \nonumber \\
    &\mathbf{s}^{l}_k = \mathtt{LayerNorm}(\mathtt{FeedForward}(\mathbf{\hat{s}}^{l}_k) + \mathbf{\hat{s}}_k^{l}). \nonumber
\end{align}
Given the final representations of the last decoder layer $\mathbf{s}_k^{N}$, the probability of current target sentence $y_k$ are computed as:
\begin{align}
p(y_k|y_{<k},x_{k},x_{-k}) &= \prod_t p(y_{k,t}|y_{k,\le t},y_{<k},x_{k},x_{-k}) \nonumber \\
&= \prod_t \mathtt{softmax}(E[y_{k,t}]^\top s_{k,t}^{N}). \nonumber
\end{align}

\section{Experiment}

We experiment on four widely used document-level parallel datasets in two language pairs for machine translation:
\begin{compactitem}

    \item TED (\textsc{Zh-En}/\textsc{En-De}). The Chinese-English and English-German TED datasets are from IWSLT 2015 and 2017 evaluation campaigns respectively. 
    We mainly explore and develop our approach on TED \textsc{Zh-En}, where we take dev2010 as development set and tst2010-2013 as testset. For TED \textsc{En-De}, we use tst2016-2017 as our testset and the rest as development set. 
    \item News (\textsc{En-De}). We take News Commentary v11 as our training set. The WMT newstest2015 and newstest2016 are used for development and testsets respectively.
    \item Europarl (\textsc{En-De}). The corpus are extracted from the Europarl v7 according to the method mentioned in \newcite{maruf2019selective}.\footnote{The last two corpora are from \newcite{maruf2019selective}}
    
\end{compactitem}

We applied byte pair encoding~\cite[BPE]{Sennrich2016Neural} to segment all sentences with 32K merge operations. We splited each document by 20 sentences to alleviate memory consumption in the training of our proposed models.
We used the Transformer architecture as our sentence-level, context-agnostic baseline and develop our proposed model on the top of it. For models on TED \textsc{Zh-En}, we used a configuration smaller than \texttt{transformer{\_}base}~\cite{Vaswani2017Attention} with model dimension $d_z=256$, dimension $d_{\rm ffn} = 512$ and number of layers $N=4$. As for models on the rest datasets, we change the dimensions to 512/2048. 
We used the Adam optimizer~\cite{KingmaB14:adam:iclr} and the same learning rate schedule strategy as \cite{Vaswani2017Attention} with 8,000 warmup steps. 
The training batch consisted of approximately 2048 source tokens and 2048 target tokens. Label smoothing~\cite{szegedy2016rethinking} of value 0.1 was used for training.
For inference, we used beam search with a width of 5 with a length penalty of 0.6. The evaluation metric is BLEU \cite{papineni2002bleu}.  We did not apply checkpoint averaging~\cite{Vaswani2017Attention} on the parameters for evaluation.

\begin{table*}[t]
      \footnotesize
   \selectfont
     \centering
     \begin{tabular}{lrcc||c||cccc}
     \toprule
         \multirow{2}{*}{\textbf{Model}} & \multirow{2}{*}{ $\Delta |\vtheta|$} &  \multirow{2}{*}{$v_{\rm{train}}$} &  \multirow{2}{*}{ $v_{\rm{test}}$}   & \multicolumn{1}{c||}{\textsc{Zh-En}} & \multicolumn{4}{c}{\textsc{En-De}} \\
          &&&& TED  & \sc TED & News &  Europarl & avg. \\
      
      \hline
         
      \textsc{SentNmt}~\cite{Vaswani2017Attention}   &   0.0m   &1.0$\times$& 1.0$\times$      & 17.0  & 23.10& 22.40 & 29.40 & 24.96 \\
      \hline
      DocT~\cite{zhang2018improving}       &9.5m&0.65$\times$&   0.98$\times$    & n/a & 24.00 & 23.08  & 29.32 & 25.46 \\
      HAN \cite{miculicich2018document} &4.8m&0.32$\times$&  0.89$\times$     & 17.9 & 24.58&\bf 25.03 & 28.60 & 26.07 \\
      SAN \cite{maruf2019selective}         &4.2m&0.51$\times$&  0.86$\times$     & n/a &24.42 & 24.84 & 29.75 & 26.33\\
      QCN~\cite{yang2019enhancing}                    &n/a&n/a&  n/a     & n/a &\bf 25.19  & 22.37 & 29.82 & 25.79 \\
      \hline
      \textsc{Ours}                                   &4.7m &0.22$\times$ &1.08$\times$       & \bf 19.1 & 25.10& 24.91 & \bf 30.40 & \bf 26.80 \\
      \hline

     \end{tabular}
    \caption{Experiment results of our model in comparison with several baselines, including increments of the number of parameters over Transformer baseline ($\Delta |\vtheta|$), training/testing speeds ($v_{\rm{train}}$/$v_{\rm{test}}$, some of them are derived from \protect\newcite{maruf2019selective}), and translation results of the testsets in BLEU score. }
    \label{tab:main}
 \end{table*}

\subsection{Main Results}
\paragraph{Document-level Translation.}
We list results of experiments in Table \ref{tab:main},  comparing four context-aware NMT models: Document-aware Transformer~\cite[DocT]{zhang2018improving}, Hierarchical Attention NMT \cite[HAN]{miculicich2018document}, Selective Attention NMT \cite[SAN]{maruf2019selective} and Query-guided Capsule Network~\cite[QCN]{yang2019enhancing}.
As shown in Table \ref{tab:main}, by leveraging document context, our proposed model obtains 2.1, 2.0, 2.5, and 1.0 gains over sentence-level Transformer baselines in terms of BLEU score on TED \textsc{Zh-En}, TED \textsc{En-De}, News and Europarl datasets, respectively. 
Among them, our model archives new state-of-the-art results on TED \textsc{Zh-En} and Europarl, showing the superiority of exploiting the whole document context. Though our model is not the best on TED \textsc{En-De} and News tasks, it is still comparable with QCN and HAN and achieves the best average performance on English-German benchmarks by at least 0.47 BLEU score over the best previous model. 
We suggest this could probably because we did not apply the two-stage training scheme used in \newcite{miculicich2018document} or regularizations introduced in \newcite{yang2019enhancing}.
In addition, while sacrificing training speed, the parameter increment and decoding speed could be manageable.


\begin{table}[t]
     \centering
      \footnotesize
  \begin{tabular}{lc}
  \toprule
      {\bf Model}                                            & Test    \\
      \hline
      \sc SentNMT                & 17.0 \\
      \hline
      {\sc DocNMT} (documents as input/output)    &14.2 \\
      HAN~\cite{miculicich2018document} & 15.6 \\
      \hline
      \textsc{Ours}                                           & 17.8 \\
      \hline
  \end{tabular}
  \caption{Results of sentence-level translation on TED \textsc{Zh-En}. }
  \label{tab:sent}
\end{table}


\paragraph{Sentence-level Translation.}

We compare the performance on single sentence translation in Table \ref{tab:sent}, which demonstrates the good compatibility of our proposed model on both document and sentence translation, whereas the performance of other approaches greatly leg behind the sentence-level baseline. 
The reason is while our proposed model does not, the previous approaches require document context as a separate input stream. This difference ensures the feasibility in both document and sentence-level translation in this unified framework.
Therefore, our proposed model can be directly used in general translation tasks with any input text of any number of sentences, which is more deployment-friendly.

\subsection{Analysis and Discussion}

\paragraph{Does Bilingual Context Really Matter? Yes.}

To investigate how important the bilingual context is and corresponding contributions of each component, we summary the ablation study in Table \ref{tab:ablation}. 
First of all, using the entire document as input and output directly cannot even generate document translation with the same number of sentences as source document, which is much worse than sentence-level baseline and our model in terms of document-level BLEU.
For source context modeling, only casting the whole source document as an input sequence (Doc2Sent) does not work.
Meanwhile, reset word positions and introducing segment embedding for each sentence alleviate this problem, which verifies one of our motivations that we should focus more on local sentences.
Moreover, the gains by the segment-level relative attention and gated context fusion mechanism demonstrate retrieving and integrating source global context are useful for document translation.
As for target context, employing Transformer-XL decoder to exploit target historically global context also leads to better performance on document translation.
This is somewhat contrasted to \cite{zhang2018improving} claiming that using target context leading to error propagation. 
In the end, by jointly modeling both source and target contexts, our final model can obtain the best performance.

\begin{table}[t]
     \centering
     \footnotesize
  \begin{tabular}{lr}
  \toprule
      {\bf Model}                                           & BLEU (BLEU$_{\rm doc}$)    \\
      \hline
      \textsc{SentNmt}~\cite{Vaswani2017Attention}                &11.4 (21.0)  \\
      {\sc DocNmt} (documents as input/output)    & n/a (17.0) \\
      \hline
      \multicolumn{2}{c}{\textit{Modeling source context}} \\
      Doc2Sent & 6.8 \\
      + reset word positions for each sentence & 10.0 \\
      + segment embedding & 10.5 \\
      + segment-level relative attention & 12.2 \\
      + context fusion gate                         & 12.4 \\
      \hline
      \multicolumn{2}{c}{\textit{Modeling target context}} \\
      Transformer-XL decoder [Sent2Doc] &  12.4\\
      \hline
      Final model [\textsc{Ours}] &  12.9 (24.4)\\
      \hline
  \end{tabular}
  \caption{Ablation study on modeling context on TED \textsc{Zh-En} development set. ''Doc" means using a entire document as a sequence for input or output. BLEU$_{\rm doc}$ indicates the document-level BLEU score calculated on the concatenation of all output sentences.}
  \label{tab:ablation}
\end{table}



\begin{figure}[t]
  \centering
  \includegraphics[width=0.43\textwidth]{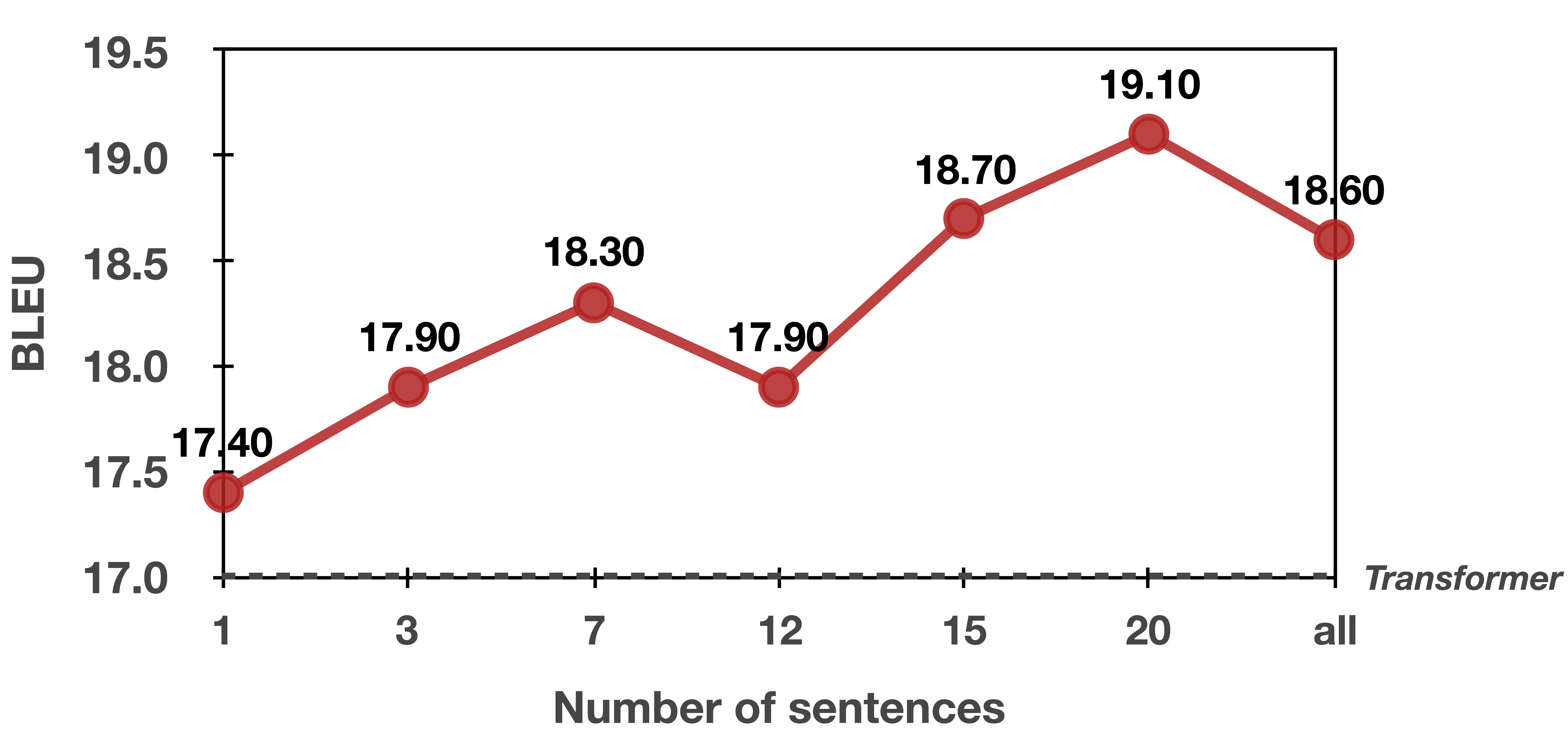}
  \caption{BLEU score w.r.t. \#sent. of context on TED \textsc{Zh-En}.}
  \label{fig:context-length}
\end{figure}

\paragraph{Effect of Quantity of Context: the More, the Better.} 
We also experiment to show how the quantity of context affects our model in document translation.
As shown in Figure \ref{fig:context-length}, we find that providing only one adjacent sentence as context helps performance on document translation, but that the more context is given, the better the translation quality is, although there does seem to be an upper limit of 20 sentences. Successfully incorporating context of this size is something related work has not successfully achieved~\cite{zhang2018improving,miculicich2018document,yang2019enhancing}. 
We attribute this advantage to our hierarchical model design which leads to more gains than pains from the increasingly noisy global context guided by the well-formed, uncorrupted local context.

\paragraph{Effect of Transfer Learning: Data Hungry Remains a Problem for Document-level Translation.}
Due to the limitation of document-level parallel data, exploiting sentence-level parallel corpora or monolingual document-level corpora draws more attention.
We investigate transfer learning (TL) approaches on TED \textsc{Zh-En}. 
We pretrain our model on WMT18 \textsc{Zh-En} sentence-level parallel corpus with 7m sentence pairs, where every single sentence is regarded as a document. We then continue to finetune the pretrained model on TED \textsc{Zh-En} document-level parallel data (source \& target TL). We also compare to a variant only whose encoder is initialized (source TL).
As shown in Table \ref{tab:TL}, transfer learning approach can help alleviate the need for document level data in source  and target languages to some extent. However, the scarcity of document-level parallel data still prevents the document-level NMT from extending at scale.

\begin{table}[t]
     \centering
      \footnotesize
  \begin{tabular}{lcc}
  \toprule
      {\bf Model}                                           & Dev  & Test  \\
      \hline
      Transformer~\cite{Vaswani2017Attention}                &11.4  & 17.0 \\
      BERT+MLM~\cite{Li2019pretrained} & n/a & 20.7 \\
      \hline
      \textsc{Ours} & 12.9 & 19.1 \\
    \textsc{Ours} + source TL                 & 13.9  & 19.7 \\
    \textsc{Ours} + source \& target TL     & 14.9 & 21.3 \\
      \hline
  \end{tabular}
  \caption{Effect of transfer learning (TL). }
  \label{tab:TL}
\end{table}

\begin{figure}[t]
  \centering
  \includegraphics[width=0.36\textwidth]{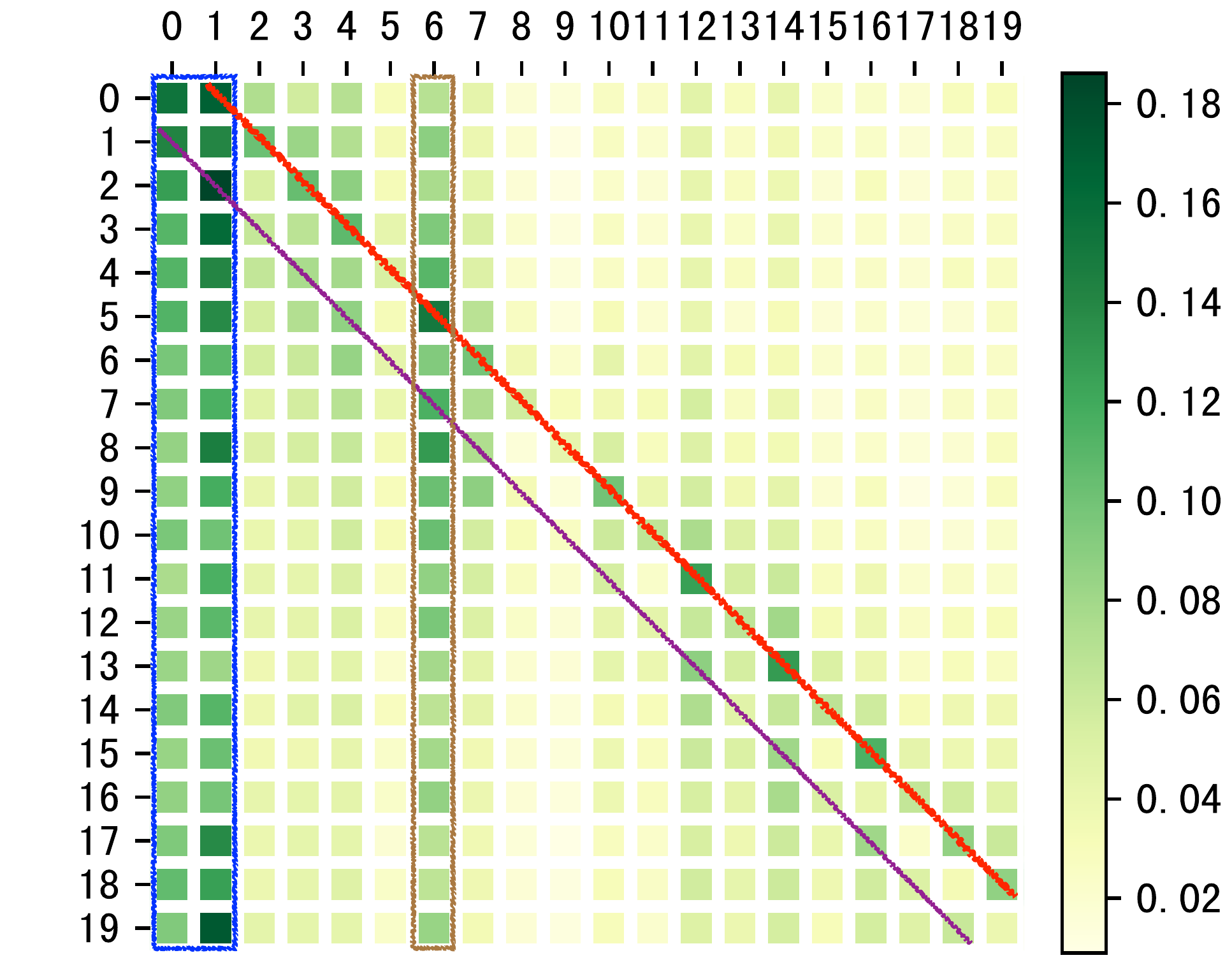}
  \caption{Visualization of sentence-to-sentence attention based on segment-level relative attention. Each row represents a sentence while each column represents another sentence to be attended. The weights of each row sum to 1.}
  \label{fig:attn}
\end{figure}

\paragraph{What Does Model Learns about Context? A Case Study.}
Furthermore, we are interested in what the proposed model learns about context. 
In Figure \ref{fig:attn}, we visualize the sentence-to-sentence attention weights of a source document based on segment-level relative attention.
Formally, the weight of the $k$-th sentence attending to the $\kappa$-th sentence are computed by $\alpha_{k}^{\kappa} = \frac{1}{|x_{k}|} \sum_i \sum_j \alpha^{\kappa,j}_{k,i}$, where $\alpha^{\kappa,j}_{k,i}$ is defined by Eq.(\ref{eq:seg-attn}). As shown in Figure \ref{fig:attn}, we find very interesting patterns (which are also prevalent in other cases): 
1) first two sentences (blue frame), which contain the main topic and idea of a document, seem to be a very useful context for all sentences; 
2) the previous and subsequent adjacent sentences (red and purple diagonals, respectively) draw dense attention, which indicates the importance of surrounding context; 
3) although sounding contexts are crucial, the subsequent sentence significantly outweighs the previous one. This may imply that the lack of target future information but the availability of the past information in the decoder forces the encoder to retrieve more knowledge about the next sentence than the previous one;
4) the model seems to not that care about the current sentence. Probably because the local context can flow through the context fusion gate, the segment-level relative attention just focuses on fetching useful global context;
5) the 6-th sentence also gets attraction by all the others (brown frame), which may play a special role in the inspected document.



\begin{table}[t]
     \centering
      \footnotesize
  \begin{tabular}{lcccc}
  \toprule
      {\bf Model}     & deixis  &    lex.c. &  ell.infl. & ell.VP  \\
      \hline
      \sc SentNmt      &50.0  &45.9 &52.2 &24.2   \\
      \sc Ours          &61.3  &46.1 &61.0 &35.6  \\
      \hline
      \newcite{voita2019good}$*$ & 81.6 & 58.1 & 72.2 &80.0 \\
      \hline
  \end{tabular}
  \caption{Accuracy ($\%$) of discourse phenomena. $*$ different data and system conditions, only for reference. }
  \label{tab:discourse}
\end{table}


\paragraph{Analysis on Discourse Phenomena.}
We also want to examine whether the proposed model actually learns to utilize document context to resolve discourse inconsistencies that context-agnostic models cannot handle. We use contrastive test sets for the evaluation of discourse phenomena for English-Russian by \newcite{voita2019good}. There are four test sets in the suite regarding deixis, lexicon consistency, ellipsis (inflection), and ellipsis (verb phrase). Each testset contains groups of contrastive examples consisting of a positive translation with correct discourse phenomenon and negative translations with incorrect phenomena. The goal is to figure out if a model is more likely to generate a correct translation compared to the incorrect variation.
We summarize the results in Table \ref{tab:discourse}. Our model is better at resolving discourse consistencies compared to context-agnostic baseline. \newcite{voita2019good} use a context-agnostic baseline, trained on $4\times$ larger data, to generate first-pass drafts, and perform post-processings, which is not directly comparable, but would be easily incorporated with our model to achieve better results.



\section{Conclusion}
In this paper, we propose a unified local and global NMT framework, which can successfully exploit context regardless of how many sentence(s) are in the input. 
Extensive experimentation and analysis show that our model has indeed learned to leverage a larger context. In future work we will investigate the feasibility of extending our approach to other document-level NLP tasks, e.g., summarization.

\section*{Acknowledgements}
Shujian Huang is the corresponding author. This work was supported by the National Science Foundation of China (No. U1836221, 61772261, 61672277). Zaixiang Zheng was also supported by China Scholarship Council (No. 201906190162). Alexandra Birch was supported by the European Union’s Horizon 2020 research and innovation programme under grant agreements No 825299 (GoURMET) and also by the UK EPSRC fellowship grant EP/S001271/1 (MTStretch).

\bibliographystyle{named}
\bibliography{ijcai20}

\end{document}